\documentclass[letterpaper, 10 pt, journal, twoside]{IEEEtran}
\ifCLASSINFOpdf
\else
\fi
 \usepackage{graphicx}
 \usepackage{amsmath} 
 \usepackage{empheq}
\usepackage{xcolor}
\usepackage{mathrsfs}
 \usepackage{amsfonts,amssymb}
 \usepackage[linesnumbered,ruled,vlined]{algorithm2e} 
\usepackage{multirow}
\usepackage{pifont} 
\usepackage{array}
\usepackage{hyperref}
\usepackage{graphicx}
\usepackage{multirow}
\usepackage{makecell}
\usepackage{amsfonts} 
\usepackage{tikz}
\usepackage{xcolor}

\newcommand{\redbox}[1]{%
    \tikz[baseline=(X.base)]{%
        \node[draw=red, thick, rectangle, inner sep=2pt] (X) {$#1$};
    }%
}
\SetKwRepeat{Do}{do}{while}%

\begin{document}
\title{Reasoning and Learning a Perceptual Metric for Self-Training of Reflective Objects in Bin-Picking with a Low-cost Camera}

% \author{Michael~Shell,~\IEEEmembership{Member,~IEEE,}
%         John~Doe,~\IEEEmembership{Fellow,~OSA,}
%         and~Jane~Doe,~\IEEEmembership{Life~Fellow,~IEEE}% <-this % stops a space
% \thanks{M. Shell was with the Department
% of Electrical and Computer Engineering, Georgia Institute of Technology, Atlanta,
% GA, 30332 USA e-mail: (see http://www.michaelshell.org/contact.html).}% <-this % stops a space
% \thanks{J. Doe and J. Doe are with Anonymous University.}% <-this % stops a space
% \thanks{Manuscript received April 19, 2005; revised August 26, 2015.}}
\author{Peiyuan Ni$^{1}$, Chee Meng Chew$^{1}$, Marcelo H. Ang Jr.$^{1}$, Gregory S. Chirikjian$^{1,2}$%
\thanks{Manuscript received: March, 25, 2025; Revised: May, 24, 2025; Accepted: July, 28, 2025. This paper was recommended for publication by Editor Valada Abhinav upon evaluation of the Associate Editor and Reviewers' comments.
This work was supported by the National Research Foundation, Singapore, under its Medium Sized Centre Programme - Centre for Advanced Robotics
Technology Innovation (CARTIN), subaward A-0009428-08-00, and AME Programmatic Fund Project MARIO (A-0008449-01-00). (\textit{Corresponding author: Gregory S. Chirikjian}) } %Use only for final RAL version
\thanks{$^{1}$The authors are with the Department of Mechanical Engineering, National University of Singapore, Singapore. pyni@nus.edu.sg or pyni\_sjtu@qq.com; \{chewcm, mpeangh, mpegre\}@nus.edu.sg}% 
\thanks{$^{2}$Gregory S. Chirikjian is  with the Department of
Mechanical Engineering, University of Delaware, Newark, DE 19716, USA.
gchirik@udel.edu}%
\thanks{Digital Object Identifier (DOI): see top of this page.}
}

\markboth{IEEE Robotics and Automation Letters. Preprint Version. AUGUST, 2025}
{NI \MakeLowercase{\textit{et al.}}: Self-Training of
Reflective Objects in Bin-Picking with a Low-cost Camera} 




\maketitle

\begin{abstract}
Bin-picking of metal objects based on low-cost RGBD cameras may suffer errors due to sparse depth information and reflective part texture, leading to a need for manual labeling. To reduce the need for human intervention, we propose a framework consisting of a metric learning stage and a self-training stage. Specifically, to automate the handling of the data captured by a low-cost camera (LC), a multi-object pose reasoning (MoPR) algorithm is proposed which optimizes the data with depth, collision and boundary constraints. To improve the accuracy of the retrieved pseudo labels, we generalize the traditional Bayesian Gaussian Mixture Model (BGMM) to the configuration manifolds and propose Symmetry-aware Lie-group based BGMM (SaL-BGMM). Moreover, to solve the ambiguity issue of perceptual metric learning for multiple objects, we propose a weighted ranking information noise contrastive estimation (WR-InfoNCE) loss. Experiments show that our algorithm outperforms related popular works both on ROBI and our proposed Self-ROBI dataset. 
\end{abstract}

% \begin{IEEEkeywords}
% IEEE, IEEEtran, journal, \LaTeX, paper, template.
% \end{IEEEkeywords}

% \fi
\IEEEpeerreviewmaketitle

\section{Introduction}
% for IEEE journal papers produced under \LaTeX\ using
% IEEEtran.cls version 1.8b and later.
\IEEEPARstart{F}{or} bin-picking of reflective objects, traditional consumer-grade cameras can be an economical and easily accessible choice compared to high-performance cameras. However, this approach may suffer from several challenges when using traditional object pose estimation algorithms: 1) sparse depth information; 2) variable and reflective object texture. Using real data can improve these while manual labeling is quite time-consuming. Therefore, it is meaningful to conduct self-training for reflective object pose estimation based on a low-cost camera (LC). Although several works \cite{Sim-to-real}\cite{Nerf-supervision} on self-training for reflective object pose estimation have been proposed, there are still two critical unsolved issues: 1) They are all based on analyzing individual objects without considering the relationships among all the objects holistically. 2) They require repetitive training iterations, object reconstruction, or isolating individual objects during self-training stage, which is not easy to use and may disrupt the target scene.
 
\begin{figure}  
    \centering
    \includegraphics[width=0.95\linewidth]{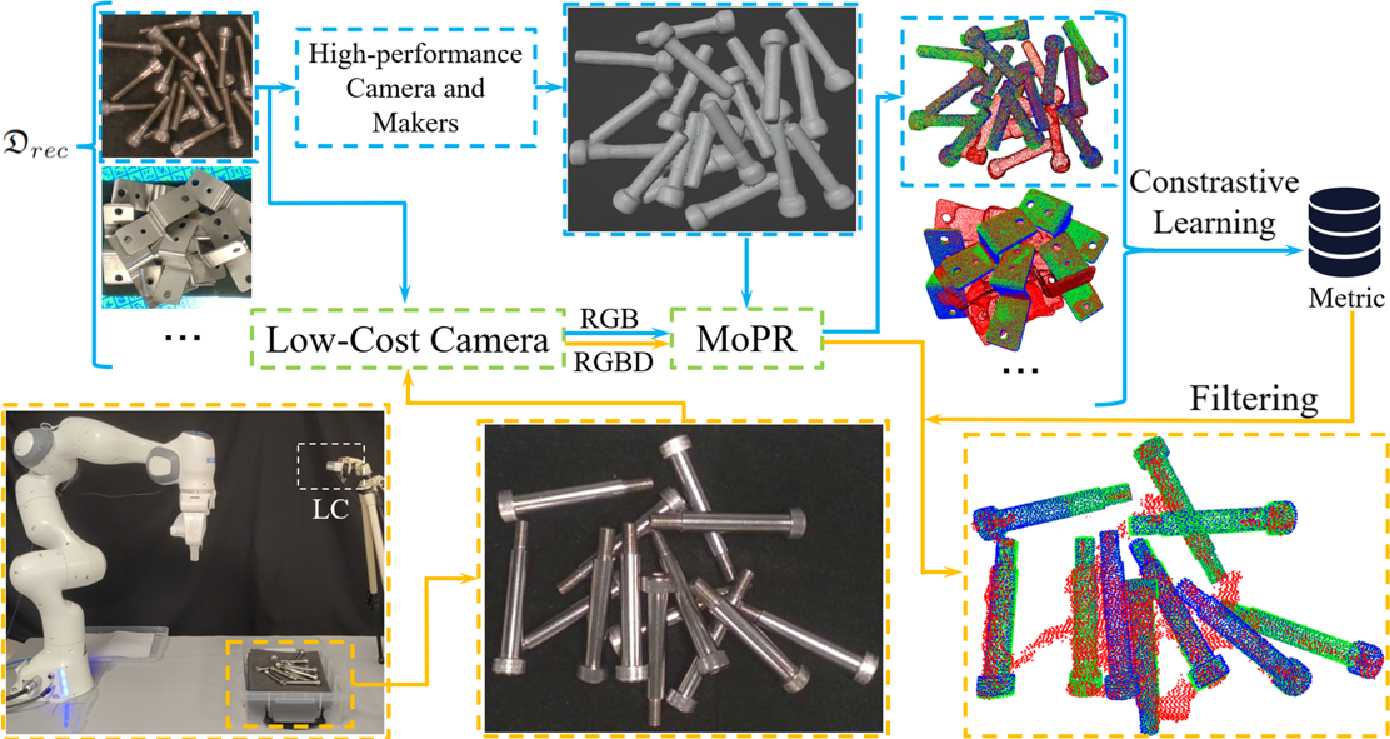}
    \caption{Overall framework. The blue and yellow parts respectively represent the processes of the metric learning and self-training stages. The green part denotes both of them. The green and  blue poses respectively denote ground truths and the estimated pseudo poses. The red point clouds denote reconstructed data or data captured by the LC. }
    \label{fig:overall}
       \vspace{-5pt}
\end{figure}

To address these challenges, we propose a metric learning stage prior to self-training. During this stage, our algorithm will learn a holistic pose metric using object models reconstructed from data
captured by the LC. The training for the metric uses models reconstructed from data obtained by a high-performance camera (HC) or QR codes as the ground truth. Then self-training can be conducted directly on real-world data or even unseen objects with only a LC. Since the configuration manifold for each object is not Euclidean space and there are ambiguities in the perceptual metric for multiple objects, we propose SaL-BGMM and WR-InfoNCE to address these issues. We summarize our contributions as follows: 1) We propose a two-stage framework to reduce the need for labeling reflective objects in the self-training stage based on an LC. 2) We present a MoPR framework based on SaL-BGMM which utilizes occlusion boundaries as cues to filter candidates. 3) We present a perceptual metric specifically designed for multiple cluttered metal objects based on a WR-InfoNCE. 4) We construct a dataset comprising household and industrial objects to validate the performance of self-training based on a Franka robot\footnote{\url{https://github.com/ChirikjianLab/Self-ROBI}}.

\section{RELATED WORK}

\subsection{Self-training for reflective object pose estimation}

Recent popular self-supervised object pose estimation often relies on texture learning \cite{selfpose++}, multi-view pseudo flow \cite{flow} or domain adaption from simulation to real scene \cite{Pseudopose}. However, they are not suitable for reflective objects as the textures are not only changeable with the viewing angles but also suffer from low signal-to-noise ratio and image saturation \cite{nextbest}, which makes these popular methods difficult to generalize to such scenarios. For self-training of reflective objects, X. Li et al. 
 \cite{Sim-to-real2} only use synthetic data to train but it still suffers from this problem. K. Chen et al. \cite{Sim-to-real} propose an iterative self-training framework to facilitate cost-effective robotic grasping but with a general perceptual metric \cite{generalmetric}. On the other hand, Y.-C. Lin et al. propose a Neural Radiance Field (NeRF) representation \cite{Nerf-supervision} of a scene to train object descriptors. T. Tan et al. \cite{ONDApose} propose an occlusion-aware neural domain adaptation method via the CAD-like radiance field. These NeRF-based methods show good performance but they work best for isolated objects. Therefore, they may not be suitable for bin-picking because they regard occlusion as noise and the texture of single object differs from the texture in cluttered environments. Additionally, recent works \cite{ZS6D}\cite{Gigapose}\cite{foundationpose} on zero-shot learning also require no manual labeling for unseen objects and can make predictions directly, but there are two issues: 1) Some works \cite{Gigapose}\cite{foundationpose} rely on rendered data, which often fail to capture real-world reflective textures. 2) Paper \cite{ZS6D} relies on a general metric, which is not designed for reflective objects and LC.

\subsection{Contrastive learning for object pose estimation}

Contrastive learning has widely been used on object pose estimation with discrete form \cite{templatepose}. C. Zhao et al.\cite{unseen} apply a continuous contrastive loss with geodesic distance as weights but it can only be applied to the rotation prediction task. B. Wen et al. \cite{foundationpose} propose pose-conditioned triplet loss for novel object tracking. All of these works mainly focus on single objects and do not compare any of two non-ground truth poses contrastively, which hinders the discrimination between different non-ground truth poses.  On the other hand,  V. N. Nguyen et al. \cite{Gigapose} propose a local contrastive learning algorithm to predict 2D-2D correspondences. Rather than predicting correspondences, we consider predicting a metric value for multiple objects by improving work \cite{ranking}.

\subsection{Gaussian Mixture Models based on Lie theory}
Gaussian Mixture Models (GMMs) can intuitively provide a good fit for nonlinear space but cannot be directly used in a configuration manifold as its metric differs from the metric in Euclidean space. Zeestraten et al. \cite{gmrlie} propose an extension for GMMs with inference that can work on Riemannian manifolds. S. Calinon et al. \cite{gmmlie} show when GMMs are applied directly to data in non-Euclidean manifolds, the clustering performance tends to be poor. And they present a method to vectorize the manifold around each cluster center. W. Liu et al. \cite{liulie} propose to use the derivatives of SE(3)  to directly optimize GMMs in SE(3) without using Lie-group-based gaussian distributions. Existing related algorithms have not addressed the clustering or fitting of object poses, which is a common application in pose estimation, so we introduce \cite{gmmlie} into BGMM with symmetry-awareness to enhance the fitting capabilities of the Cross Entropy Method \cite{cem}. 
  
\section{Method}

\subsection{Overall Frameworks} 
Shown in Fig. \ref{fig:overall}, the whole framework is divided into two parts: metric learning stage and self-training stage. The whole targets have two goals: 1) No human labeling is required for both of two stages. 2) The user or the consumer can directly perform self-training in a single view of a LC without any other dependencies (e.g., markers or a HC) or disrupting the target scene. The LC we use is a traditional consumer-grade camera, which is not ideal for capturing or reconstructing metallic scenes.

Similar to \cite{generalmetric}, the learned perceptual metric in our work aims to evaluate the pose state in self-training stage. In the metric learning stage, our approach uses a HC or other localization devices to reconstruct the scene. In our paper, we choose a high-resolution camera and markers to obtain the reconstructed data, based on Geo-neus \cite{Geo-neus}. At the same time, the LC will also capture the data with the same scene. With the help of Markers, it is easy to register the reconstructed data with the data captured by LC. Then both of them are sent to the MoPR part. MoPR performs joint optimization for multiple objects, which will be introduced in Part B. Then the optimal poses are regarded as pseudo labels for metric learning and applied to train a perceptual metric, which will be introduced in Part C.
In self-training stage, given objects unseen before, which shares the same background, light and camera position as test stage, we directly use the data (low-resolution RGB and sparse depth data) captured by LC and send them to MoPR part. Finally, the learned perceptual metric is applied to filter the candidates to obtain the best pseudo labels for self-training.
 
\subsection{Multi-object pose reasoning}
  
\begin{figure}[thpb]
    \centering
    \includegraphics[width=1\linewidth]{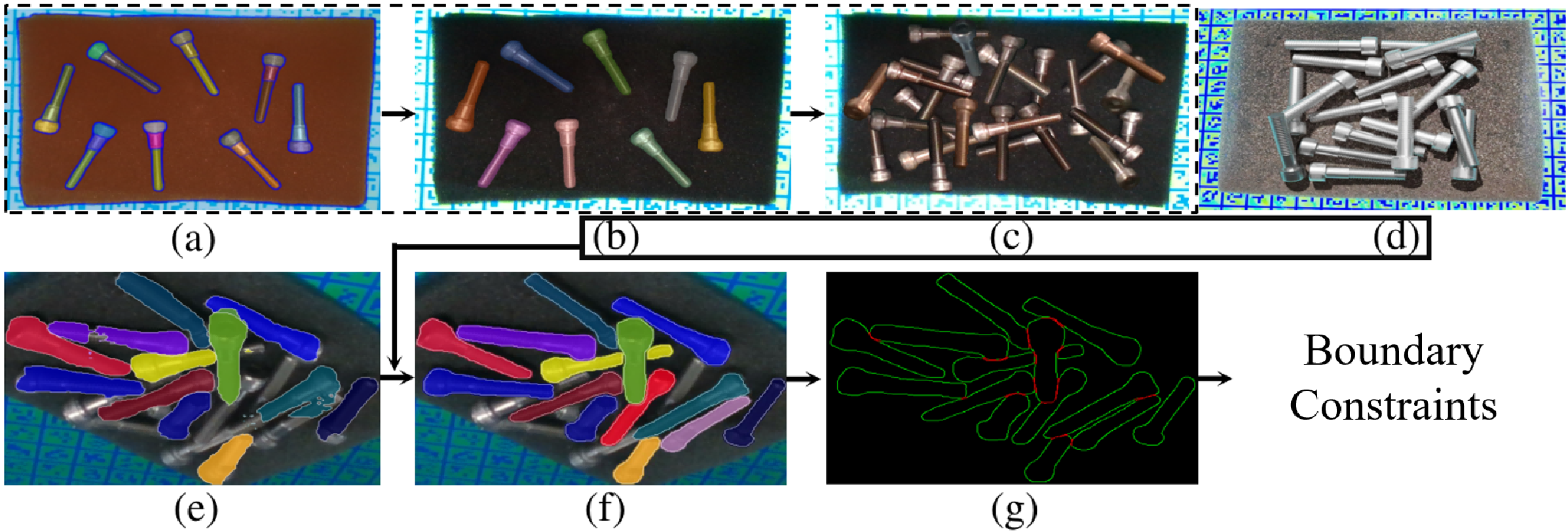}
    \caption{Self-training for object segmentation: (a) Segmentation results from SAM; (b) Post-processing results; (c) Augmented data; (d) Synthetic data; (e) Results trained only by synthetic data; (f) Final segmentation results; (g) Extracted boundaries.}
    \label{fig:frameworks}
\end{figure} 

For object segmentation, we first place objects at some intervals, as shown in Fig. \ref{fig:frameworks} (a), to ensure that the Segment Anything Model (SAM) \cite{SAM} can segment the objects robustly. Since SAM tends to segment subparts for each object, we merge the masks that belong to the same connected non-background component (Fig. \ref{fig:frameworks} (b)). Then a copy-paste \cite{Copypaste} augmentation is applied to generate the Augmented data for bin-picking (Fig. \ref{fig:frameworks} (c)). Moreover, we also render the simulation data with domain randomization (Fig. \ref{fig:frameworks} (d)). Then we use the data from Fig. \ref{fig:frameworks} (b)$\sim$(d) to train the object segmentation algorithm \cite{maskrcnn}, shown in Fig. \ref{fig:frameworks} (f). After that, we extract the edges (Fig. \ref{fig:frameworks} (g)) and convert it into boundary constraints that will be introduced in next section.

\subsubsection{\textbf{Constraints definition}} 
We consider three constraints to optimize our problem: collision, depth and boundary constraints. Given objects $i$ and $j$ with poses $\textbf{T}_i$ and $\textbf{T}_j$, let $\textbf{g}^{i}$ denote the array of occupancy grids \cite{collision} when the object is in pose $\textbf{T}_i$. Let $\mathcal{L}_{\textbf{C}_{ij}}$ denote the collision loss for $\textbf{T}_i$ and $\textbf{T}_j$. We set $\mathcal{L}_{\textbf{C}_{ij}}=\frac{sum\left( \textbf{g}^{i} \circ \textbf{g}^{j} \right)}{sum \left(\textbf{g}^{i}\right)}$, where $\circ$ denotes element-wise product. $sum$() sums all the elements for the input array. For depth constraint, given object $i$, we select the Chamfer distance (CD) proposed in \cite{Sim-to-real} (Eq. 5 from paper \cite{Sim-to-real}) within the boundary between input depth data and rendered point clouds, denoted by $D_{i}$. Let $\mathcal{L}_{\textbf{D}_{i}}$ denote the depth loss for $\textbf{T}_i$ and we set $\mathcal{L}_{\textbf{D}_{i}}=D_{i}$. For boundary constraints, we consider two cases: \textbf{Case 1} (Constraint between each two objects) and \textbf{Case 2} (Constraint for each single object). For \textbf{Case 1}, we denote the overlapping boundaries between object $i$ and $j$ extracted from SAM (Fig. \ref{fig:frameworks} (g)) by $\textbf{O}_s$. The non-overlapping boundaries near $\textbf{O}_s$ within a threshold are also considered and the extended boundaries including both overlapping and non-overlapping parts are denoted by $\textbf{E}_s$. For rendered edges with poses $\textbf{T}_i$ and $\textbf{T}_j$, these two types of boundaries are denoted as $\textbf{O}_r$ and $\textbf{E}_r$, shown in Fig. \ref{fig:edges} (a). Let $\mathcal{L}_{\textbf{B}_{ij}}$ denote the loss for matching performance of \textbf{Case 1} with $\textbf{T}_i$ and $\textbf{T}_j$:
  
\begin{figure} 
    \centering
    \includegraphics[width=1\linewidth]{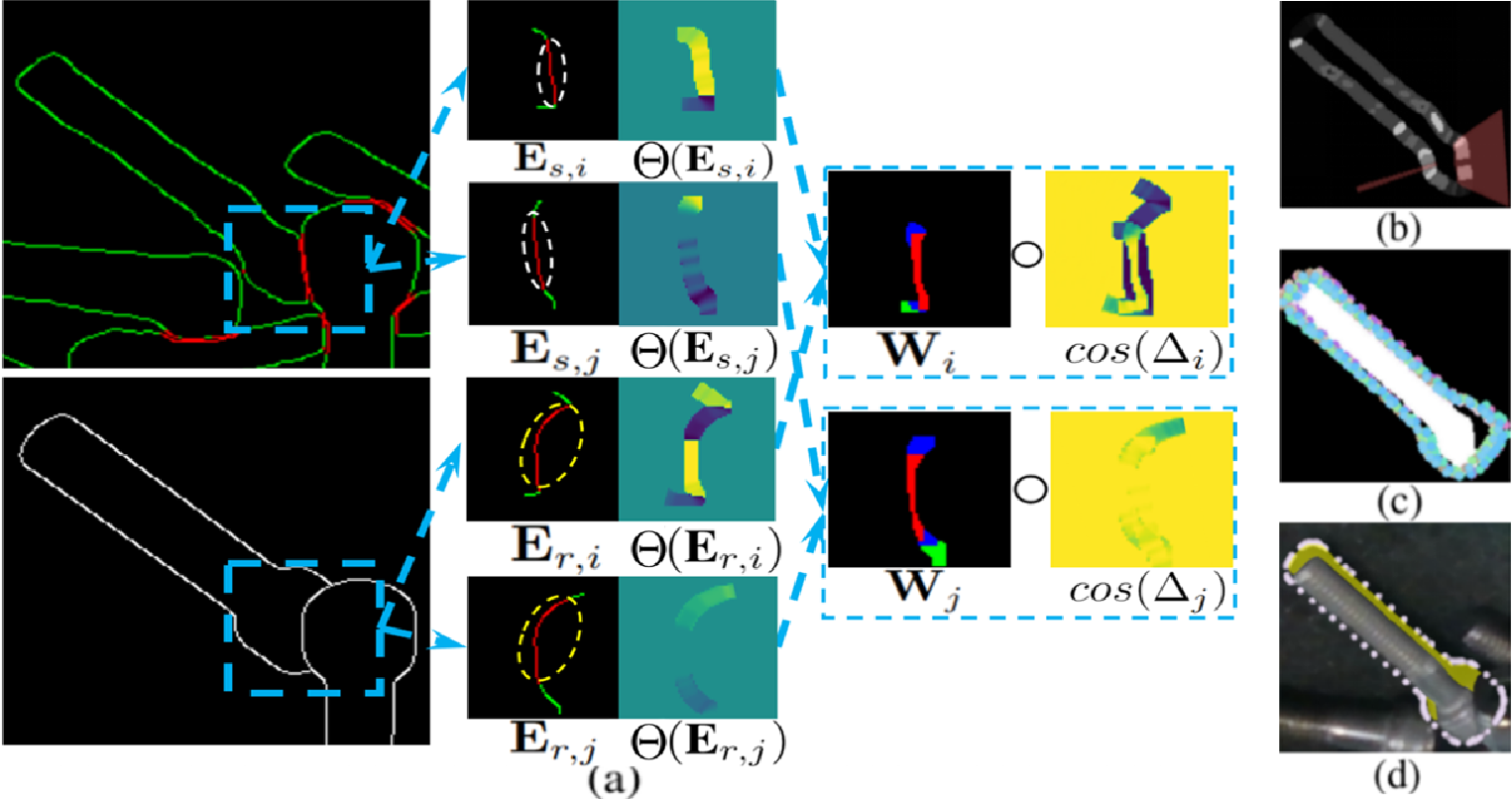}
    \caption{(a) Illustration of Eq. \ref{boundary}. The red and green edges in  $\textbf{E}_s$ and $\textbf{E}_r$ are overlapping and non-overlapping boundaries. The edges in white dashed circles are $\textbf{O}_s$ and the edges in yellow dashed circles are $\textbf{O}_r$. The red, green and blue areas on $\textbf{W}_i$ and $\textbf{W}_j$ respectively indicate the regions of $\textbf{M}_1$, $\textbf{M}_2$ and $\textbf{M}_3$. (b) Occlusion mask for spread gradients of object $i$. (c) Top 50 candidates, $R_{lm}$ of which are larger than 90. (d) Ground truth pose, $R_{lm}$ of which equals 71. The yellow parts are the unrecognized mask.}
    \label{fig:edges}
       \vspace{-5pt}
\end{figure}

\begin{equation}
\label{boundary}
\mathcal{L}_{\textbf{B}_{ij}}=- \frac{1}{2}
\sum_{k\in \lbrace i,j \rbrace }\frac{sum(\textbf{W}_k \circ cos(\Delta_{k} ) )}{sum(\mathcal{M}(\textbf{E}_{s,k}))}\text{,}  
\end{equation}
where $\Delta_{k}=\Theta(\textbf{E}_{s,k})-\Theta(\textbf{E}_{r,k})$. $\Theta$() is an operation to apply Gaussian blurring with a kernel size of (7,7) to input edges, and then extract their image gradient orientation. The orientation at each pixel ranges from $0$ to $2\pi$. $\mathcal{M}$() denotes the operation to apply the same Gaussian blurring to input edges and extract their mask by binarizing it to 0 and 1. $\textbf{W}_k$ = $\lambda_1\textbf{M}_1$ + $\lambda_2\textbf{M}_2$ + $\lambda_3\textbf{M}_3$. $\lambda_t$ ($t=1, 2, 3$) is the weight of $\mathbf{M}_t$ ($t=1, 2, 3$). $\textbf{M}_1$, $\textbf{M}_2$ and $\textbf{M}_3$ are shown as follows:
\begin{equation}
 \textbf{M}_1=\mathcal{M}(\textbf{O}_{s,k})\circ \mathcal{M}(\textbf{O}_{r,k})
\end{equation}
\begin{equation}
\textbf{M}_2=(\mathcal{M}(\textbf{E}_{s,k})-\mathcal{M}(\textbf{O}_{s,k}))\circ (\mathcal{M}(\textbf{E}_{r,k})-\mathcal{M}(\textbf{O}_{r,k}))
\end{equation}
\begin{equation}
\textbf{M}_3=\mathcal{M}(\textbf{E}_{s,k})\circ \mathcal{M}(\textbf{E}_{r,k})-\textbf{M}_1-\textbf{M}_2
\end{equation} 

$\circ$ denotes the element-wise product. $\textbf{O}_{s,k}$ is $\textbf{O}_s$ belonging to the $k$-th object and other similar variables follow the same principle. As shown in Fig. \ref{fig:edges} (a), $\textbf{M}_1$ and $\textbf{M}_2$ respectively indicate the matched spatial areas for overlapping and non-overlapping parts between extracted edges and rendered edges, so we assign high values to the weights
(i.e., $\lambda_1$ and $\lambda_2$) of $\textbf{M}_1$ and $\textbf{M}_2$. In $\textbf{M}_3$, the extracted and rendered edges are matched, but the matched categories (overlapping and non-overlapping parts) are incorrect, indicating a region with high uncertainty. We assign it with a low weight, i.e., $\lambda_3$. In our experiment, $\lambda_1$, $\lambda_2$ and $\lambda_3$ are set to 1, 1 and 0.3.
 
For \textbf{Case 2}, we use score of Line2D (Eq. 2 from paper \cite{linemod}), denoted by $R_{lm}$$\in$[0,100], to evaluate the matching result of $i$th object under pose $\textbf{T}_i$ with its corresponding boundaries. As Line2D \cite{linemod} ignores the occlusion problem, we should construct a mask to cover the occluded spread gradients generated by Line2D  that does not belong to object $i$, thereby improving the matching accuracy, as shown in the red mask of Fig. \ref{fig:edges} (b). However, this requires determining whether object $i$ at $\textbf{O}_s$ is being occluded or is occluding object $j$. Specifically, if the state of the overlapping boundary $\textbf{O}_s$ is occluding object $j$, which means $\textbf{O}_s$ is the object's contour, we do not need to mask the spread gradient generated by $\textbf{O}_s$; otherwise, we need to mask it. We define the occlusion state of object $i$ relative to object $j$ at $\textbf{O}_s$ as a hidden variable $u_{i,j}\in\lbrace 0,1\rbrace$, where 0 and 1 denote occluding and occluded states. We can also get $u_{i,j}+u_{j,i}=1$. We define all the occlusion states of object $i$ as vector $\textbf{u}_i$. The loss for \textbf{Case 2} is denoted by $\mathcal{L}_{\textbf{L}_{i}}$ and we set $\mathcal{L}_{\textbf{L}_{i}}=-R_{lm}$, which is a function of $\textbf{T}_i$ and $\textbf{u}_i$. 
 
\subsubsection{\textbf{SaL-BGMM}}
\begin{figure}
 
    \centering
\includegraphics[width=0.7    \linewidth]{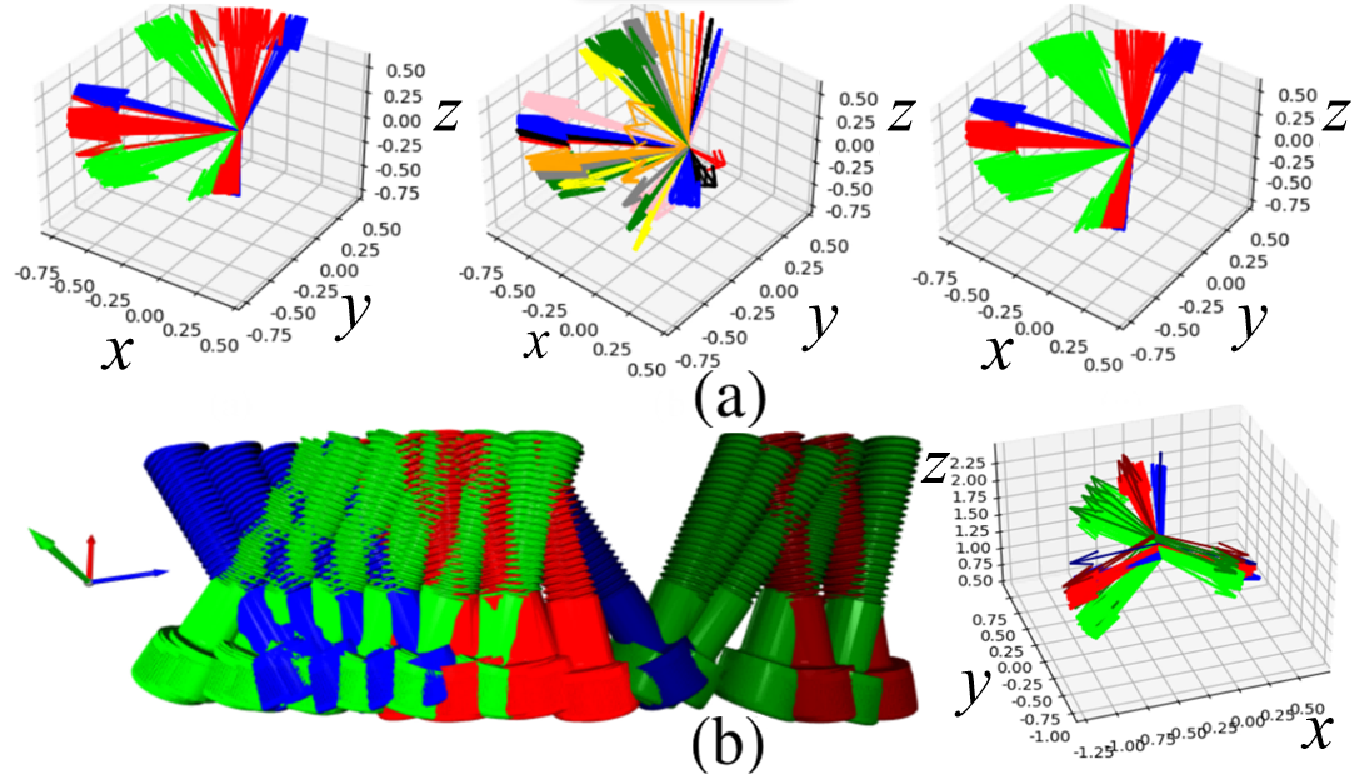}
\caption{(a) SaL-BGMM  for SO(3). Left: Ours without Lie group. Mid: Ours without symmetry-awareness. Right: Full algorithm. (b) SaL-BGMM for PCG(3), where colors with the same hue but different shades represent elements belonging to the same SO(3) space but different $\mathbb{R}^3$ spaces.} 

    \label{fig:BGMM}
       \vspace{-5pt}
\end{figure}

As not all the constraints in part $\textit{1)}$ are analytical, we should consider a non-differentiable method to optimize them. We choose multi-extremal Cross Entropy Method (CEM) \cite{cem} and the key to this problem is how to fit the sampled candidates to generate the multi-extremal elite candidates. Usually, BGMM is a good option as it only requires the input of maximum number of clusters, denoted by $N_{c}$. Therefore, we should generalize BGMM from Euclidean space to configuration manifold. Firstly, we use the pose change group, denoted by PCG(3) \cite{pgc}, to represent the pose, which is defined by a direct product of the rotation and translation groups, denoted by SO(3)$\times\mathbb{R}^3$. As SO(3) and $\mathbb{R}^3$ are independent of each other, we first perform clustering on rotation and then on translation. Given $\boldsymbol{x},\boldsymbol{y}$$\in$SO(3), let $\mathcal{T}_{\boldsymbol{y}}$ denote tangent space at $\boldsymbol{y}$. We define $\mathrm{Log}_{\boldsymbol{y}}\boldsymbol{x}$ to transfer $\boldsymbol{x}$ into $\mathcal{T}_{\boldsymbol{y}}$ by $\log(\boldsymbol{y}^{-1}\boldsymbol{x})$, which is based on the local frame at $\boldsymbol{y}$. Similarly, given tangent vector $\boldsymbol{\tau}$ from $\mathcal{T}_{\boldsymbol{y}}$, we define $\mathrm{Exp}_{\boldsymbol{y}}\boldsymbol{\tau}$ to transfer $\boldsymbol{\tau}$ into SO(3) by $\boldsymbol{y}\exp(\boldsymbol{\tau})$. Then we consider a Lie group based gaussian distribution \cite{gmmlie} and make sure the computation of the Euclidean space be restricted to the tangent spaces of clusters' means. Specifically, we transform part of the original equations of BGMM (Eq. 10.52, Eq. 10.53, Eq. 10.61, Eq. 10.62, Eq. 10.64 in \cite{prml})  as follows:

\begin{equation}\label{q1}
\overline{\mathbf{x}}_k = \frac{1}{N_k} \sum_{n=1}^{N} r_{nk} \redbox{\mathrm{Log}_{m_k'}(\mathbf{x}_n)} 
\end{equation}
\begin{equation}\label{q2}
 S_k = \frac{\sum_{n=1}^{N} r_{nk} ( \redbox{\mathrm{Log}_{m_k}(\mathbf{x}_n) - \tilde{\mathbf{x}}_k})(\redbox{\mathrm{Log}_{m_k}(\mathbf{x}_n) - \tilde{\mathbf{x}}_k})^T}{N_k}
\end{equation}
\begin{equation}\label{q3}
m_k =\redbox{\mathrm{Exp}_{m_k'}}(\frac{1}{\beta_k} (\beta_0 \redbox{\mathrm{Log}_{m_k'}(m_0)} + N_k \overline{\mathbf{x}}_k))
\end{equation} 
  
\begin{equation}\label{q4}
W_k^{-1} =\tilde{W} + \frac{\beta_0 N_k (\redbox{\tilde{x}_k - \mathrm{Log}_{m_k}{m_0}})(\redbox{\tilde{x}_k - \mathrm{Log}_{m_k}{m_0}})^T}{\beta_0 + N_k} 
\end{equation} 
\begin{eqnarray}\label{q5} 
\mathbb{E}_{\mu_k, \Lambda_k} \left[ (x_n - \mu_k)^T \Lambda_k (x_n - \mu_k) \right] \nonumber\\
= D \beta_k^{-1} + \nu_k \redbox{\mathrm{Log}_{m_k}(\mathbf{x}_n)^T} \mathbf{W}_k \redbox{\mathrm{Log}_{m_k}(\mathbf{x}_n)}
\end{eqnarray}  
\textit{The variables in Eq. \ref{q1}$\sim$Eq. \ref{q5} are independent of the other variables in this paper, which follow the definitions of BGMM (Eq. 10.49 $\sim$ Eq. 10.66 in \cite{prml})}, where $\tilde{W}= W_0^{-1} + N_k S_k$ and $ \tilde{\mathbf{x}}_k=\mathrm{Log}_{m_k}(\mathrm{Exp}_{m_k'}(\overline{\mathbf{x}}_k))$. $m_k'$ denotes $m_k$ from the previous iteration. To facilitate readers' understanding, we have highlighted the differences between Eq. \ref{q1}$\sim$Eq. \ref{q5} and Eq. 10.49 $\sim$ Eq. 10.66 in \cite{prml} with red boxes. To remove symmetry ambiguity for all the rotation poses, we run Lie group-based BGMM twice. For the first running, the mean with max component is recorded, denoted by $\textbf{R}_0$$\in$SO(3). Then for each remaining pose, denoted by $\textbf{R}_t$$\in$SO(3), it will be updated as follows combined with geodesic distance:
\begin{equation}\label{sym}
\hat{\textbf{R}}_t= \left(\underset{\textbf{S} \in \mathcal{S}(\textbf{R}_t)}{{\arg\min} \,}{\rm arccos}\left(\frac{tr((\textbf{S}\textbf{R}_t)^T\textbf{R}_0)-1}{2}\right)\right)\textbf{R}_t\text{,}
\end{equation}
where $\hat{\textbf{R}}_t$ denotes the updated $\textbf{R}_t$ and $\mathcal{S}(\textbf{R}_t)$ denotes  the set of all symmetric transformations of $\textbf{R}_t$. If the symmetric transformation of $\textbf{R}_t$ is continuous, such as axial symmetry, it will be discretized. Then all the $\hat{\textbf{R}}_t$ will be selected for the second running. Finally, for each cluster of SO(3), we run traditional BGMM for its $\mathbb{R}^3$ space, shown in Fig. \ref{fig:BGMM} (b).
 
\subsubsection{\textbf{Non-differentiable pose optimization}} 
As illustrated in part $\textit{1)}$, for all the losses, we regard them as energy functions and convert them into likelihoods by Gibbs distribution \cite{prml}. Let $\textbf{X}$ denote the observed data. Taking $\mathcal{L}_{\textbf{C}_{ij}}$ for an example, the likelihood for collision constraint of  $\textbf{T}_i$ and $\textbf{T}_j$ (denoted by p$_{\textbf{C}}$($\textbf{X}\vert\textbf{T}_i,\textbf{T}_j$)) is calculated by p$_{\textbf{C}}$($\textbf{X}\vert\textbf{T}_i,\textbf{T}_j$) = $\exp(-\gamma\mathcal{L}_{\textbf{C}_{ij}})$/$Z$, where $Z$ is a  normalization factor and can be ignored for negative log-likelihood (NLL) calculation. We use the same operations for other losses ($\mathcal{L}_{\textbf{B}_{ij}}$, $\mathcal{L}_{\textbf{D}_{i}}$ and $\mathcal{L}_{\textbf{L}_{i}}$). The values of $\gamma$ for $\mathcal{L}_{\textbf{C}_{ij}}$, $\mathcal{L}_{\textbf{B}_{ij}}$, $\mathcal{L}_{\textbf{D}_{i}}$ and $\mathcal{L}_{\textbf{L}_{i}}$ are set to 10$^2$, 1, 10$^3$ and 0.01. We accumulate their likelihoods for the whole objects and denote them by \text{p}$_{\textbf{C}}$(\textbf{X}$\vert$\textbf{T}), \text{p}$_{\textbf{B}}$(\textbf{X}$\vert$\textbf{T}), \text{p}$_{\textbf{D}}$(\textbf{X}$\vert$\textbf{T}) and \text{p}$_{\textbf{L}}$(\textbf{X}$\vert$\textbf{T}, \textbf{u}). The whole likelihood can be formulated as:
\begin{eqnarray}\label{nll}
\text{p}(\textbf{X}|\textbf{T},\textbf{u})=\text{p}_{\textbf{C}}(\textbf{X}|\textbf{T})\text{p}_{\textbf{B}}(\textbf{X}|\textbf{T})\text{p}_{\textbf{D}}(\textbf{X}|\textbf{T})\text{p}_{\textbf{L}}(\textbf{X}|\textbf{T},\textbf{u}) 
\end{eqnarray}
In the EM algorithm \cite{prml}, minimizing the NLL (i.e., $-ln$\{\text{p}(\textbf{X}$\vert$\textbf{T},\textbf{u})\}) can be converted into maximizing the Evidence Lower Bound (ELBO). Let $E$ denote the ELBO.
\begin{align}\label{elbo}
    & E=\sum_{\textbf{u}}\text{p}(\textbf{u}|\textbf{X};\textbf{T}')ln\{\text{p}(\textbf{X},\textbf{u}|\textbf{T})\} \nonumber\\
    &=ln\{\text{p}_{\textbf{C}}(\textbf{X}|\textbf{T})\text{p}_{\textbf{B}}(\textbf{X}|\textbf{T})\text{p}_{\textbf{D}}(\textbf{X}|\textbf{T})\} \nonumber \\
    &+\underbrace{\sum_{i}\sum_{\textbf{u}_i}\text{p}(\textbf{u}_i|\textbf{X};\textbf{T}')ln\{\text{p}_{\textbf{L}}(\textbf{X}|\textbf{u}_i,\textbf{T}_i)\text{p}(\textbf{u}_i)\}}_{\textbf{L}^*}
\end{align}
 The second term is denoted by $\textbf{L}^*$. $\textbf{T}'$ denotes $\textbf{T}$ in the previous iteration. $\text{p}(\textbf{u}_i)$ is similar to the mixing coefficients in GMMs and follows a multivariate Bernoulli distribution. Specifically, we have $\text{p}(\textbf{u}_i) = \prod_{j} \text{p}(u_{i,j})$, where $j$ belongs to the set of indices of objects neighboring object $i$. Due to the influence of brightness, there are errors for extracted boundaries of metallic objects as shown in Fig. \ref{fig:edges} (d). Although these errors will not affect the rough shape of the extracted boundaries, they do impact the precise positioning of the objects. It can also be observed that $R_{lm}$ of ground truth pose is much lower than  top 50 $R_{lm}$ of the whole candidates (Fig. \ref{fig:edges} (c)) although their rotations are close. Therefore, using $\textbf{L}^*$ directly is inaccurate but $\textbf{L}^*$ can provide good constraints on the shape of the object. So we use a softened ELBO, denoted by $\tilde{E}$, to replace Eq.\ref{elbo}, where $\textbf{L}^*$ is converted into a soft bounding constraint:
\begin{equation}\label{finalelbo}
\tilde{E}=ln\{\text{p}_{\textbf{C}}(\textbf{X}|\textbf{T})\text{p}_{\textbf{B}}(\textbf{X}|\textbf{T})\text{p}_{\textbf{D}}(\textbf{X}|\textbf{T})\}  
\end{equation}
$Subject\ to:$
$$
\textbf{R}_i\in\mathcal{B}(\{\textbf{R}_i^*\})
$$
$$
\|\mathcal{P}(\textbf{t}_i)-\mathcal{P}(\textbf{t}_i+\underset{\boldsymbol{\Delta}\in\mathcal{B}(\{\textbf{t}_i^*\})}{{\arg\min} \,}\|\textbf{t}_i-\boldsymbol{\Delta}\|_2 )\|_2< \Delta_{thr}
$$
 
Let $\{\textbf{T}_i^*\}$ denote the set of top $K$ poses obtained from $\textbf{L}^*$ for object $i$. $\{\textbf{R}_i^*\}$ and $\textbf{R}_i$ denote the rotation parts of $\{\textbf{T}_i^*\}$ and $\textbf{T}_i$, while $\{\textbf{t}_i^*\}$ and $\textbf{t}_i$ denote their translation parts. $\mathcal{B}(\{\textbf{R}_i^*\})$ and $\mathcal{B}(\{\textbf{t}_i^*\})$ are bounding volumes of $\{\textbf{R}_i^*\}$ and $\{\textbf{t}_i^*\}$, which are actually the GMMs within three standard deviations from the means fitted by SaL-BGMM. $\mathcal{P}(\textbf{t}_i)$ is the pixel coordinates of $\textbf{t}_i$ projected on the RGB image. $\Delta_{thr}$ is a pixel threshold. As Eq.\ref{finalelbo} is non-differentiable to $\text{p}(\textbf{u})$, we approximate the update of $\text{p}(\textbf{u})$ using the derivative of original ELBO (Eq.\ref{elbo}) with respect to $\text{p}(\textbf{u})$. Similar to the derivation of hidden variables for GMMs \cite{prml}, $\text{p}(\textbf{u})$ can be updated shown in Eq.\ref{u}, taking $\text{p}(u_{i,j})$ as an example.  
\begin{eqnarray}\label{u}
\text{p}(u_{i,j}=0)=  \textstyle \sum _x\text{p}(u_{x,y}=t|\textbf{X};\textbf{T}')/2 \nonumber\\
= \frac{1}{2} \sum_{x}\frac{\text{p}_{\textbf{L}_x}(\textbf{X}|u_{x,y}=t;\textbf{T}')\text{p}(u_{x,y}=t;\textbf{T}')}{\sum_{k}\text{p}_{\textbf{L}_x}(\textbf{X}|u_{x,y}=k;\textbf{T}')\text{p}(u_{x,y}=k;\textbf{T}')}\text{,} 
\end{eqnarray}
where $k\in \lbrace0,1\rbrace$ and $x\in \lbrace i,j\rbrace$. If $x=i$, then $t=0$ and $y=j$; If $x=j$, then $t=1$ and $y=i$. $\text{p}(u_{x,y}=t;\textbf{T}')$ can be regarded as the $\text{p}(u_{x,y}=t)$ in the previous iteration. $\text{p}_{\textbf{L}_x}(\textbf{X}|u_{x,y}=k;\textbf{T}')$  denotes $\text{p}_{\textbf{L}}(\textbf{X}|u_{x,y}=k;\textbf{T}')$ for object $x$. $\text{p}(u_{i,j}=1)$ can be calculated by $1-\text{p}(u_{i,j}=0)$. Then MoPR is proposed in Algorithm 1.
  
For Step 2, if $t_{1}$$=$1, each element of $\text{p}(\textbf{u})$ will be initialized with 0.5. For Step 3, $\text{p}(\textbf{u}_i)$ in $\textbf{L}^*$ can be regarded as constant  to calculate $\underset{\textbf{T}}{argmax}\ \textbf{L}^*$. The calculation of $\text{p}(\textbf{u}_i|\textbf{X};\textbf{T}')$ in $\textbf{L}^*$ can refer to $\text{p}(u_{x,y}=t|\textbf{X};\textbf{T}')$ in Eq.\ref{u}. We use the template matching algorithm from Line2D \cite{linemod} to rapidly obtain potential optimal poses and then search $\{\textbf{T}^*\}$ within these poses. Steps from 4 to 8 are actually CEM \cite{cem}. For Step 8, we first employ SaL-BGMM to fit the PCG(3) space for each object. For Step 5, if $t_{2} = 1$, we just directly sample from $\{\textbf{T}^*\}$. If $t_{2} > 1$, $\{\textbf{T}^*\}$ are sampled from the fitted model based on Step 8. The sampling number is set to $K$. Moreover, each sampled pose is refined by ICP (Interative Closest Point) \cite{liulie}, where the max correspondence distance is set to 4mm. This setting ensures that ICP matching will only use points within a local range for matching, thereby obtaining more potential solutions. In Step 9, the poses with best $\tilde{E}$ are selected as $\textbf{T}'$.

\begin{algorithm}   
  \For{$t_{1}\leftarrow1\ \mathbf{to}$ N}{
    Update $\text{p}(\textbf{u})$ with Eq.\ref{u}\; 
    $\{\textbf{T}^*\}\leftarrow \underset{\textbf{T}}{argmax}\ \textbf{L}^*$\;
    \For{$t_{2}\leftarrow1\ \mathbf{to}$ M}{
    $\{\textbf{T}\}\leftarrow$ Sample with Local ICP\;
    $\{\tilde{E}\}\leftarrow$ Calculate values of Eq.\ref{finalelbo}\;
    $\{\textbf{T}\}\leftarrow$ Select $\{\textbf{T}\}$ with top $\eta$  of $\{\tilde{E}\}$\;  
    Fit $\{\textbf{T}\}$ with SaL-BGMM if $t_{2}<M$\;}
    $\textbf{T}'\leftarrow$Select the best candidate from $\{\textbf{T}\}$\;

    }
 
\caption{MoPR}
\end{algorithm}

$N$, $M$, $K$, $N_{c}$ and $\eta$ are set to 5, 5, 50, 10 and 0.3. $\Delta_{thr}$ is set to 5. Moreover, we set conservative thresholds for collision and depth losses with 0 and 5 mm. Any candidate failing to meet these thresholds will be ignored.   
\subsection{Metric learning}
MoPR is applied for both metric learning stage and self-training stage shown in Fig. \ref{fig:overall}. For metric learning stage, we only select $\textbf{T}$ with best $\tilde{E}$ from $\{\textbf{T}\}$, denoted by $\textbf{T}^{0}$, to train a perceptual metric prediction network. For self-training stage, we select all the filtered candidates $\{\textbf{T}\}$ and use the learned perceptual metric network to filter out the best pseudo labels. 

During metric learning stage, given $\textbf{T}^{0}$, both depth and normal data for the objects with poses $\textbf{T}^{0}$ are rendered and randomly
cropped together
with RGB data captured by LC. The cropped size, denoted by $S$, is set to be proportional to the maximum size of object's model. The rendered depth and normal are contacted into a tensor denoted by $\textbf{I}_r$. To match the size of $\textbf{I}_r$, the RGB data will be converted to grayscale. Then it will be sharpened, and then merged with the original grayscale image into a new tensor, denoted as $\textbf{I}_g$. Both $\textbf{I}_r$ and $\textbf{I}_g$ will be scaled to the size of 128$\times$128 and sent to a siamese network, shown in Fig. \ref{fig:metrics} (a). The specific structure of backbone is derived from the encoder part of \cite{unseen}. The outputs of the last three layers of the backbone are followed by normalization and a 1$\times$1 convolution to obtain multi-scale features. Then we calculate pixel-wise similarities between the outputs of these two streams with cosine distance and take the average of them into a single value, which serves as the perceptual metric of the crop image and is denoted by $\boldsymbol{s}_{0}$. To perform contrastive learning, we consider a random perturbation applied to $\textbf{T}^{0}$ within the crop range to generate noise poses, denoted by $\textbf{T}^{i}$, $i$=1,2,...$k'$, the perceptual metrics of which are denoted by $\boldsymbol{s}_{i}$, $i$=1,2,...$k'$. Then we introduce Ranking InfoNCE \cite{ranking} and select CD as the ranking metric. However, if the contrastive weight for each pair of samples is equal, the results will be influenced by ambiguous labels as the labels of ambiguous samples are hard to distinguish and may not even be correct. Therefore, we propose WR-InfoNCE, denoted by $\mathcal{L}_{wr}$: 
\begin{figure}
  
    \centering
    \includegraphics[width=1\linewidth]{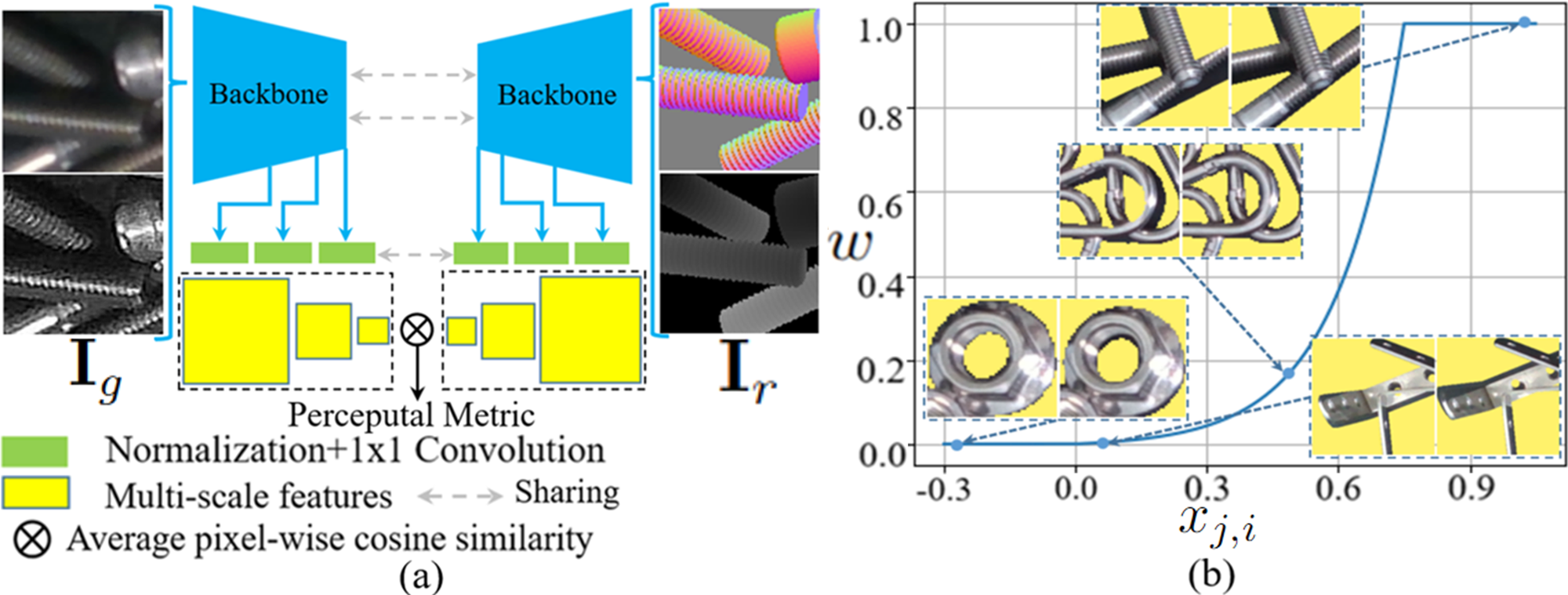}
    \caption{(a) Network structure for metric learning. (b) Curves of $w$. For better illustration of the difference between rendered poses and RGB modality, the mask of rendered normals is projected on RGB image, which is highlighted in yellow. The images on the left and right of each demo correspond to the RGB modality associated with $\textbf{T}^{j}$ and $\textbf{T}^{i}$ respectively.}
    \label{fig:metrics}
     \vspace{-5pt}
\end{figure} 
\begin{equation}\label{contrastive}
 \mathcal{L}_{wr}=-\sum_{i}\log\frac{\exp(\boldsymbol{s}_{i}/\tau)}{\sum_{j}w(x_{j,i})\exp(\boldsymbol{s}_{j}/\tau)}  
\end{equation} 
$e_{i}$ denotes the CD within the range of cropped image between rendered point clouds of $\textbf{T}^{i}$ and $\textbf{T}^{0}$. $i$, $j$ = 0,1,...$k'$. $x_{j,i}=n(e_{j})-n(e_{i})$ and $n(e_{i})$ = $ e_{i}\cdot\frac{128}{S}$, which aims to normalize the CD across different objects into a unified metric space. $w(x_{j,i})$ is a piecewise function: if $x_{j,i}<0$, $w(x_{j,i})$$=$0; if $x_{j,i}>x_{limit}$, $w(x_{j,i})$$=$1; Otherwise, $w(x_{j,i})$$=$$x_2e^{x_1(x_{j,i}-x_0)}-x_3$. The thresholds ($x_0$, $x_1$, $x_2$, $x_3$) ensure  have a smooth transition from 0 to 1 exponentially, which makes sure the loss pay less attention to data with high ambiguity. Fig. 6 (b) shows some demos of RGB modality for different augmented noised poses and their corresponding positions on curve $w(x_{j,i})$. When $x_{j,i}$ approaches 0, it becomes increasingly difficult to rank $\textbf{T}^{j}$ and $\textbf{T}^{i}$, so $w$ will also drop to 0. We train our network for 200 epochs with a batch size of 16 and an initial learning rate of $10^{-4}$ using a NVIDIA Tesla V100 GPU. In our experiment, $\tilde{e}$ and $k'$ are set to 2.5 and 32. $x_{limit}$ and $\tau$ are set to 0.75 and 0.1. $x_0$,  $x_1$, $x_2$ and $x_3$ are calculated and set to 0.5, 7, 0.2 and 0.006. $S$ is set to 800$l_{max}$+50. $l_{max}$ denotes the diameter of object's bounding sphere, with meters as the unit. 

During self-training stage, we generate cropped images using a sliding window with cropped size as the stride. For each window, learned metric will be used to do the evaluation among all the candidates $\{\textbf{T}\}$. The top 10 candidates will be voted as 10, 9,...,1,  with the remaining receiving 0. Finally, the candidates with the highest voting score are considered the best pseudo labels for self-training, shown in Fig. \ref{fig:expdemo} (a). 
  
\section{Experiment}
 
\subsection{Datasets}
\begin{figure}
    
     \centering
     \includegraphics[width=0.85 \linewidth]{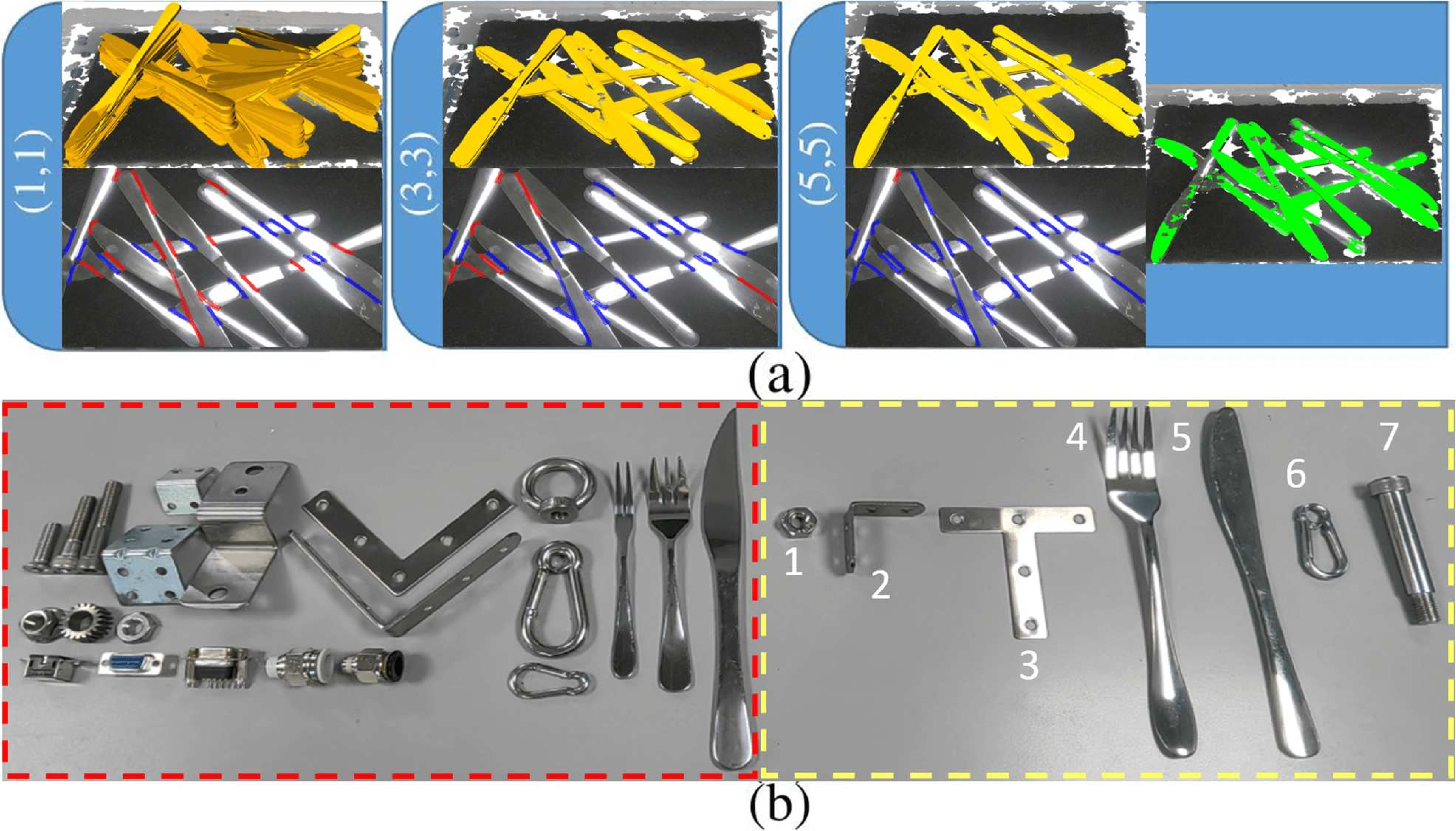}
\caption{(a) An example for MoPR and learned metric, where the red and blue edges represent misclassified and correctly classified latent variables \textbf{u} for the overlapping boundaries. The yellow and green candidates denote the sampled poses in Step 5 of MoPR and the poses filtered by learned metric. The numbers in parentheses represent $t_{1}$ and $t_{2}$ in MoPR, respectively. (b) Self-ROBI dataset: $\mathfrak{D}_{rec}$ is collected from objects in red box. $\mathfrak{D}_{self}$ and $\mathfrak{D}_{test}$ are collected from objects in yellow box.}

     \label{fig:expdemo} 
      \vspace{-5pt}
 \end{figure}

The dataset most relevant to our research is the ROBI \cite{robi}, but ROBI is primarily designed for industrial objects and consists of only seven objects, which is insufficient to form separate datasets for metric learning, self-training, and testing. Therefore, we propose a dedicated dataset to validate self-training of bin-picking for reflective objects: Self-ROBI. As shown in Fig. \ref{fig:expdemo} (b) , the whole dataset consists of three parts: 1) Metric learning part, denoted by $\mathfrak{D}_{rec}$, consisting of 600 scenes: These objects are used for learning a good perceptual metric. They consist of 22 different types of objects shown in the red box of Fig. \ref{fig:expdemo} (b). For each type of object, we collect 8$\sim$12 instances and randomly stacked them. Then we use Geo-neus and QR codes to reconstruct them. The camera used for reconstruction is Huawei Pura 70 Pro with resolution of 3840×2160. 2) Self-training part, denoted by $\mathfrak{D}_{self}$, consisting of 140 scenes: They consist of 7 different types of reflective objects for bin-picking task including 3 household objects; 3) Test part, denoted by $\mathfrak{D}_{test}$, consisting of 35 scenes: These objects belong to the same categories as those in $\mathfrak{D}_{self}$, as well as the background and light, but untrained before. The LC used for self-training and test is Realsense D415 with resolution of 1280×720. For ground truth of $\mathfrak{D}_{test}$, we place QR codes around the totebox and objects, then reconstruct the scene using Geo-neus. Then we manually label the poses with these reconstructed results. 
 
\subsection{Ablation Study for MoPR}  
We choose $20\%$ of  $\mathfrak{D}_{rec}$ to do the ablation study and show the performance of generated pseudo labels for metric learning. For $\mathfrak{D}_{rec}$, we choose the best poses generated by MoPR. We use ICP to label the ground truth and then adjust them manually. Both recall and precision of ADD-S that is less than 0.1 of the object diameter
(ADD-S(0.1$d$)) \cite{linemod}\cite{FFB6D} are selected as the evaluation metrics, where $d$ denotes object's diameter. The results are shown in Tab. 1.  
 \begin{table}[h]
 
\caption{Ablation Study for MoPR based on ADD-S(0.1$d$)}\textbf{Sym} denotes symmetry awareness (Eq.\ref{sym}). \textbf{Lie} denotes Lie group based method (Eq.\ref{q1}$\sim$Eq.\ref{q5}). \textbf{Soft} denotes softening the linemod constraint of ELBO (Eq.\ref{elbo}$\rightarrow$Eq.\ref{finalelbo}). 1 and 0 respectively denote Eq.\ref{finalelbo} and Eq.\ref{elbo} are selected as the ELBO. -1 denotes Eq.\ref{elbo} is selected but $\textbf{L}^*$ is removed. \textbf{p$_{\textbf{B}}$($\textbf{X}$$\vert$$\textbf{T}$)} denotes if we use \textbf{p$_{\textbf{B}}$($\textbf{X}$$\vert$$\textbf{T}$)} in Eq.\ref{finalelbo}. The best results are in bold.
\label{table_example}
\begin{center}
\begin{tabular}{|c|c|c|c|c|c|c| }
\hline 
 \multicolumn{1}{|c}{{ Index }}&\multicolumn{1}{|c|}{{\textbf{Soft}}}&\multicolumn{1}{c|}{{\textbf{p$_{\textbf{B}}$($\textbf{X}$$\vert$$\textbf{T}$)}}}&\multicolumn{1}{c|}{{\textbf{Lie}}}&\multicolumn{1}{c|}{{\textbf{Sym}}}&Precision&Recall \\\hline
1&1 &\checkmark&&\checkmark&91.23&83.17 \\\hline 
2&1 &\checkmark&\checkmark&&87.17&79.18 \\\hline 
3&0 &\checkmark&\checkmark&\checkmark&84.71&77.15  \\\hline 
4&-1 &\checkmark&\checkmark&\checkmark&89.83&81.74 \\\hline 
5&1 & &\checkmark&\checkmark&68.31&72.73 \\\hline
6&1 &\checkmark&\checkmark &\checkmark&\makecell{\textbf{96.51}}&\makecell{\textbf{85.25}} \\\hline

\end{tabular}
\end{center}
\end{table}

As shown in Indexes 1, 2, and 6 of Tab. 1, SaL-BGMM improves a lot compared to pure BGMM (Index 1) and Lie group based BGMM (Index 2). Pure BGMM can easily confuse different clusters for SO(3) space shown in Fig. \ref{fig:BGMM} (a), which makes it more difficult to identify potential multiple extremes with the same number of sampling points. Lie group based BGMM (Index 2) improves this but it will generates many redundant clusters that can be merged. From Indexes 3 and 6, we can find the original ELBO (Eq.\ref{elbo}) will lead to poor performance because extracted boundaries  contain noise. If $\textbf{L}^*$ is removed (Index 4), the performance is still worse than our full algorithm (Index 6) because it is easy to generate samples that exceed the soft constraint during sampling process of MoPR. From Indexes 5 and 6, we can find our proposed boundary constraint make a boost to our algorithm (68.31$\rightarrow$96.51). From the column of recall, we can also find our full algorithm (Index 6) can still achieve a recall of 85.25 while maintaining high precision. 
 
\subsection{Experience on Metrics Learning}
  
\begin{figure}
 
    \centering
\includegraphics[width=0.91 \linewidth]{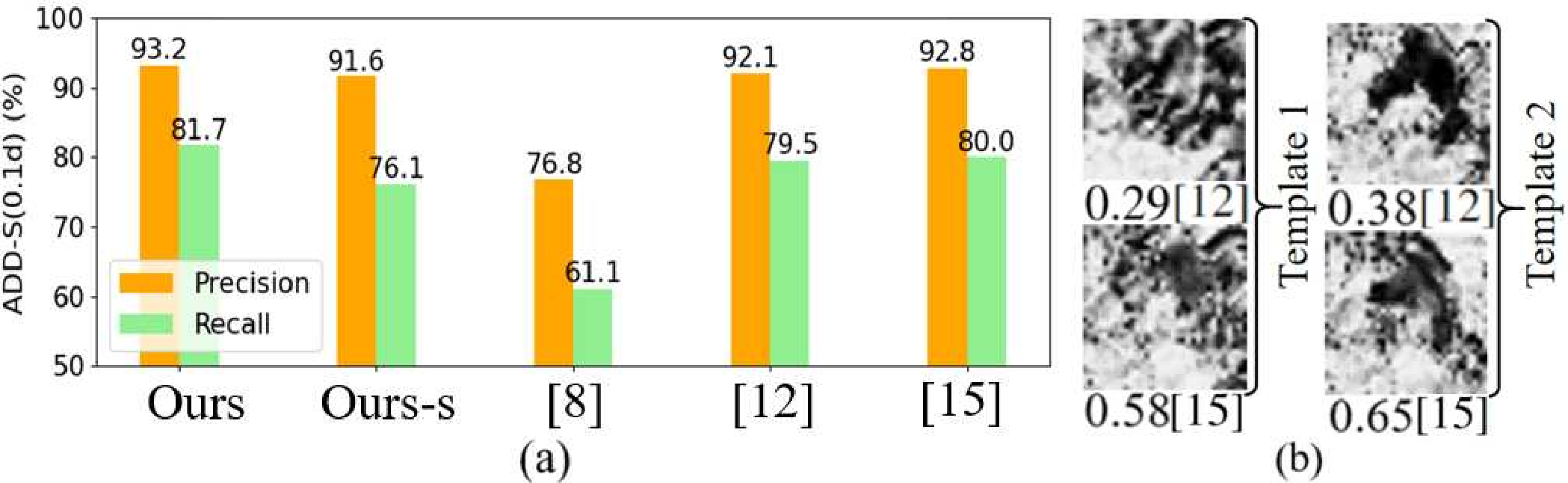}
    \caption{(a) Comparison for metric learning. Ours-s denotes our algorithm using single-object data. (b) Similarities using \cite{foundationpose} and \cite{ranking}  for Template 1 and 2 in the last row of Fig. \ref{fig:metricdemo}. }
    \label{fig:metric}
     
\end{figure}
  
\begin{figure}
  
    \centering
    \includegraphics[width=0.94  \linewidth]{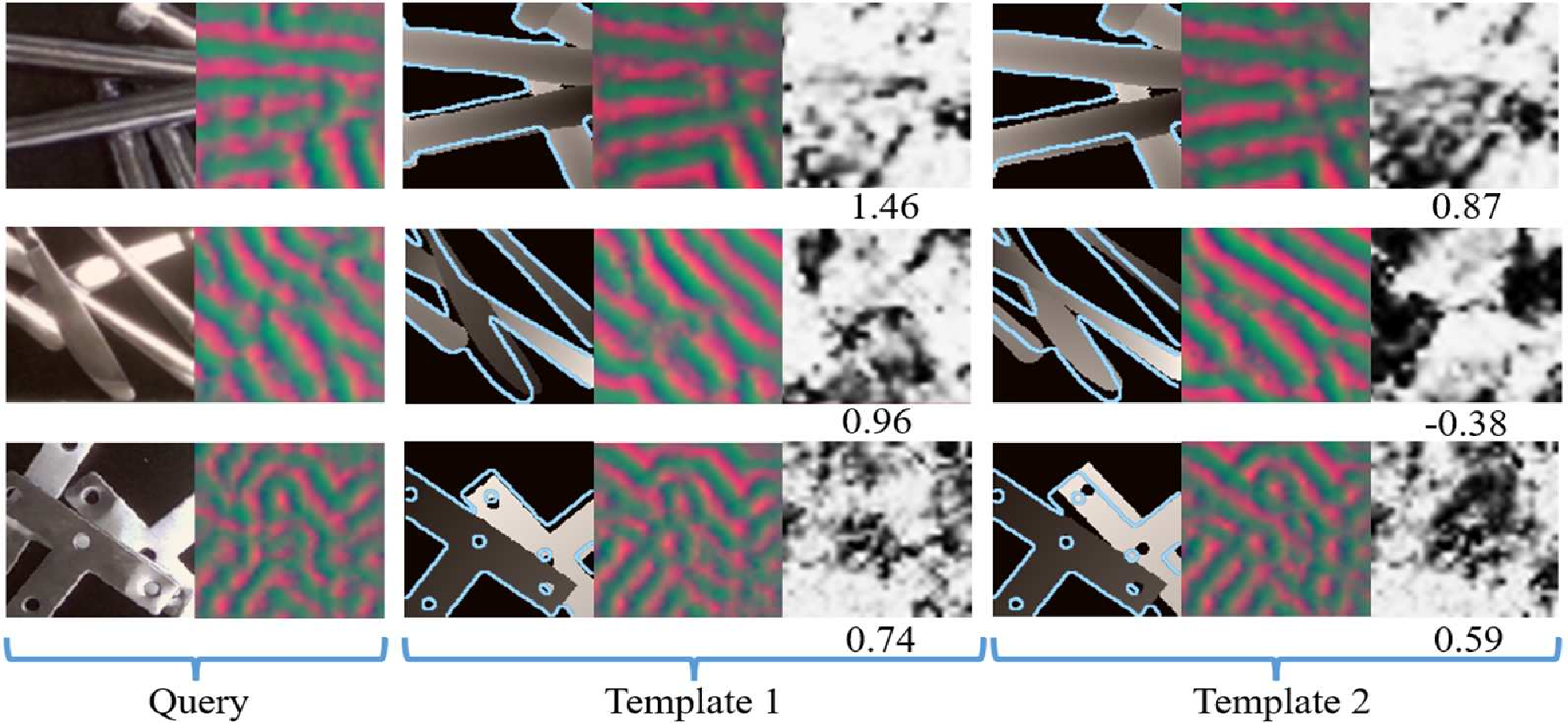}
    \caption{Visualization for learned metric, where Template 1 has a smaller error with the ground truth than Template 2. For each template ($\textbf{I}_r$), we show its depth, features and similarities with $\textbf{I}_g$. The features and similarities correspond to those at the largest scale in the multi-scale representation. The boundaries of the mask corresponding to the ground truth are projected onto the depth map. The numbers in the figure are the perceptual metrics.}
    \label{fig:metricdemo}
\end{figure}

We compare our full algorithms including WR-InfoNCE based on $\mathfrak{D}_{test}$ with three baselines: 1) General Perceptual Metric (GPM) \cite{Sim-to-real}\cite{generalmetric}; 2) Pose-conditioned Triplet Loss (PTL) \cite{foundationpose}. 3) Ranking InfoNCE \cite{ranking}. Following \cite{Sim-to-real}, we apply GPM to filter the candidate poses generated by MoPR. For PTL, we consider the candidates whose average ADD-S with $\textbf{T}^{0}$ is less than 0.1$d$ as positive. Moreover, we also conduct another comparison where we crop each object individually and use the single-object data as the metric input shown as the 'Ours-s' in Fig. \ref{fig:metric}. For 'Ours-s', the cropping center is set to the center of the object rendered by the candidate pose in RGB image. From Fig. \ref{fig:metric} (a), we can observe that using single object as the metric input leads to a decrease both in precision (93.2 $\rightarrow$91.6) and recall (81.7 $\rightarrow$76.1). This is mainly because the metric using single object faces more occlusions compared to that using multiple objects, particularly in  bin-picking problems. Moreover, GPM performs the worst as it can not generalize to multiple reflective objects. Compared to other baselines (Fig. \ref{fig:metric} (b)), our metric considers the occlusion as a cue and allows the similarity to change in a reasonable direction with respect to matching errors, which is beneficial for score voting, even if the variations of the matching errors are small, as shown in Fig. \ref{fig:metricdemo}. Because our proposed loss considers the ambiguity issue, it is less affected by fuzzy erroneous labels, whereas Ranking InfoNCE is not.

\begin{table}[h]
\caption{Pose estimation and grasping results for ROBI and Self-ROBI} Accuracy of ADD-S(0.1$d$) \cite{FFB6D} is applied to do the evaluation and the best results are in bold. The numbers in parentheses represent the grasp success rate. R1$\sim$R7 denote the objects in ROBI dataset: Zigzag, Chrome Screw, Gear, Eye Bolt, Tube Fitting, Din Connector and D-Sub Connector. S1$\sim$S7 denote the objects in $\mathfrak{D}_{test}$ of Self-ROBI shown in Fig. \ref{fig:expdemo} (b). 
\label{table_example}
\begin{center}
\begin{tabular}{|c|c|c|c|c|c|c|c|c|c|c|c|c|c|c|c|c|c|c|}
\hline
\multicolumn{1}{|c|}{ }&Fdpose\cite{foundationpose}&ONDA\cite{ONDApose}&Sim2real\cite{Sim-to-real}&Ours\\\hline
R1&43.85&54.32&47.34&\textbf{58.42}\\\hline
R2&68.42&80.50&81.58&\textbf{85.15}\\\hline
R3&91.50&95.51&\textbf{97.41}&96.78\\\hline
R4&87.37&94.20&91.13&\textbf{95.35}\\\hline
R5&86.55&\textbf{98.03}&97.72&97.22\\\hline
R6&33.83&32.01&27.18&\textbf{34.06}\\\hline
R7&25.70&25.43&23.21&\textbf{27.04}\\\hline
mean&62.46&68.76&66.51&\textbf{70.57}\\\hline
S1 & 47.54(67.95) & 54.10(72.93) & 50.82(71.10) & \textbf{55.74}(\textbf{75.32}) \\ \hline S2 & 58.14(63.73) & 67.44(77.60) & 62.79(76.62) & \textbf{69.77}(\textbf{79.21}) \\ \hline S3 & 74.42(81.88) & \textbf{83.72}(\textbf{94.58}) & 79.07(88.72) & 79.07(92.50) \\ \hline S4 & 65.85(77.14) & 70.73(85.42) & 68.29(83.20) & \textbf{73.17}(\textbf{86.20}) \\ \hline S5 & 73.81(82.92) & 76.19(89.50) & \textbf{78.57}(\textbf{92.62}) & \textbf{78.57}(91.35) \\ \hline S6 & 47.73(52.95) & 59.09(67.17) & 54.55(66.07) &   \textbf{61.36}(\textbf{68.07})\\ \hline S7 & 74.47(84.07) & 80.85(\textbf{92.23}) & 76.60(90.50) & \textbf{82.98}(91.92) \\ \hline

mean & 63.14(72.95) & 70.30(82.78) & 67.24(81.26) & \textbf{71.52}(\textbf{83.51}) \\ \hline

\end{tabular}
\end{center}
 \vspace{-5pt}
\end{table} 
\subsection{Test Results and Robot Experiment}
  
 \begin{figure}
  
     \centering \includegraphics[width=0.98 \linewidth]{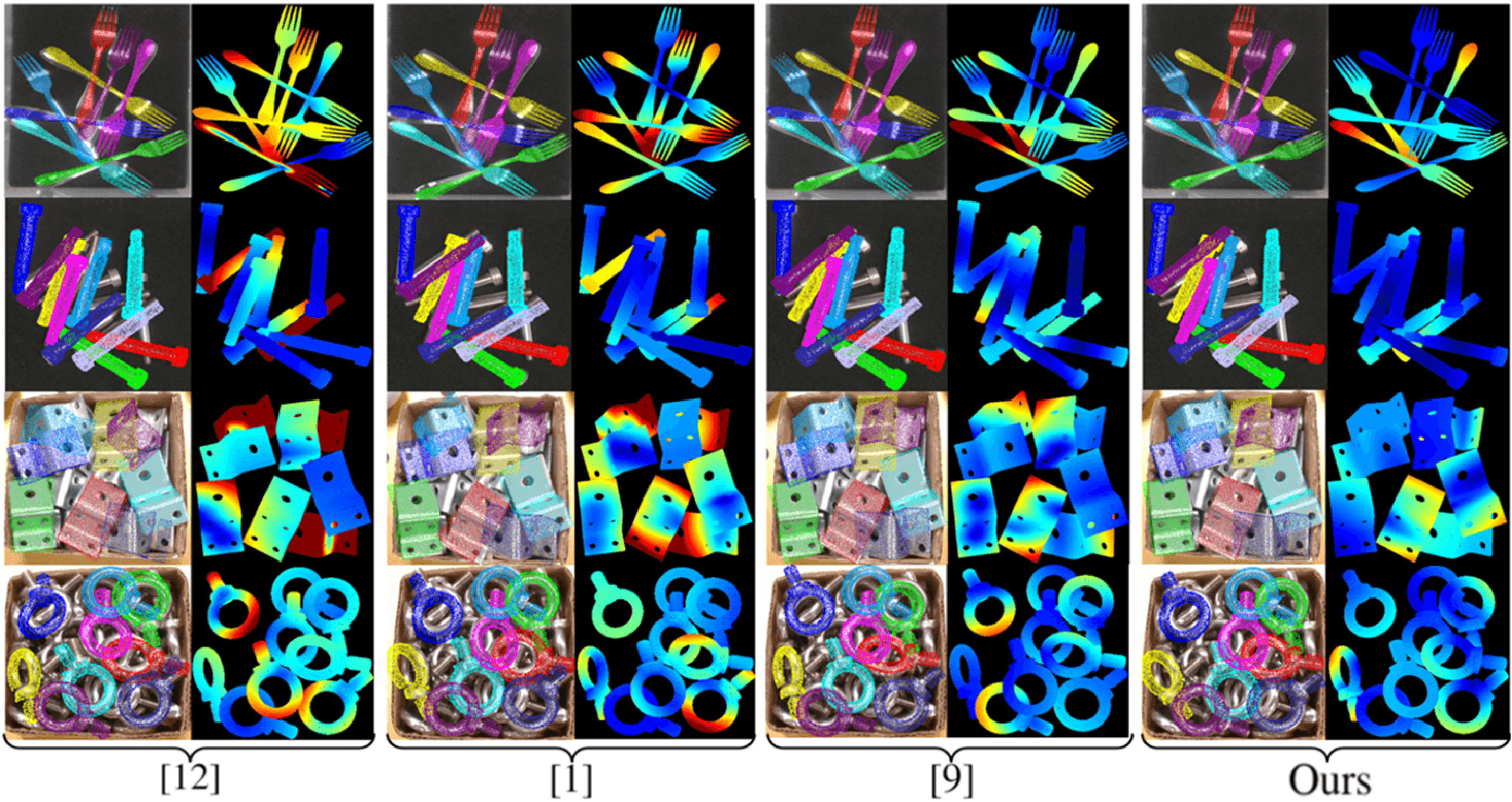}
     \caption{Test performance in Self-ROBI and ROBI. For each method, we show both predicted poses and matching errors with ground truth. Each pixel value in the colormap indicates the magnitude of the error between the ground truth and predicted poses at the corresponding pixel, with colors ranging from blue to red indicating the errors from small to large.  }
     \label{fig:Test_performance} 
      \vspace{-5pt}
 \end{figure}

We use both ROBI and Self-ROBI to test our algorithm and other baselines. Since we do not have physical objects of ROBI, we send images with sparsely distributed objects (low-bin scenarios) from ROBI into the process shown in Fig. \ref{fig:frameworks} to self-supervise the training of the segmentation algorithm. We share the segmentation results and synthetic data of bin-picking with all baselines. For ROBI, following \cite{Sim-to-real}, we select two-thirds for self-training, with the remaining parts used for evaluation. For Self-ROBI, we select $\mathfrak{D}_{self}$  for self-training and $\mathfrak{D}_{test}$ for evaluation. The backbone we selected to train pseudo labels generated by self-training is FFD6D \cite{FFB6D}. We adopt following baselines with no manual labeling to do the comparison. 1) Foundationpose (Fdpose) \cite{foundationpose}. 2) ONDA \cite{ONDApose}. 3) Sim2real \cite{Sim-to-real}. Both \cite{ONDApose} and \cite{Sim-to-real} rely on initial poses from a pose estimation algorithm trained on synthetic data. Similar to the settings of Fig. 2 (d), for each type of object from $\mathfrak{D}_{test}$, we selected 8$\sim$12 instances to build the scenario in the simulator, and we generated 1,000 synthetic scenes for each object. FFB6D \cite{FFB6D} is selected as this initial pose provider by training these synthetic data. Our method also utilizes these initial values. Specifically, in Step 3 of MoPR (Algorithm 1), we filter out the candidates  whose rotation errors with respect to the initial poses are greater than 90°. We use the accuracy of ADD-S (0.1$d$) \cite{FFB6D} to evaluate test performances in $\mathfrak{D}_{test}$ for different baselines. Moreover, we also test the grasping performances of different baselines using a Franka Robot and we evaluate them using the grasp success rate. For each type of object, we test 6 scenes. For each scene, we attempt up to 20 grasps with 6$\sim$10 objects and ensure identical conditions across all baselines. The grasp configuration is generated offline based on the object's CAD model. 

As shown in Tab. 2, Fdpose does not perform well especially for highly reflective objects (R1$\sim$R5 and S1$\sim$S7), even though it employs language-aided randomization in synthetic data. However, Fdpose shows a better performance than ONDA
and Sim2real for R6 and R7 as the complex appearance of these two objects hinders the effectiveness of ONDA and Sim2real. Our algorithm perform well both on ROBI with result of 70.57 and Self-ROBI with result of 71.52, especially for S4, S6, R1 and R2, which enables our algorithm to achieve the highest grasp success rate. Even for objects with complex appearances (R6 and R7) or large variations in scale (S1 and S4), our algorithm can still adapt to them. We also show some examples in Fig. \ref{fig:Test_performance}. We find Fdpose is worse than other methods, because it struggles to adapt to the metallic sheen and incomplete depth data. ONDA shows good performance because of its occlusion-aware domain adaptation method but it is worse than ours as it still suffer from severe occlusions and depends more heavily on the initial values of FFB6D \cite{FFB6D} than ours.   
   
\begin{figure}  
 
    \centering
    \includegraphics[width=1\linewidth]{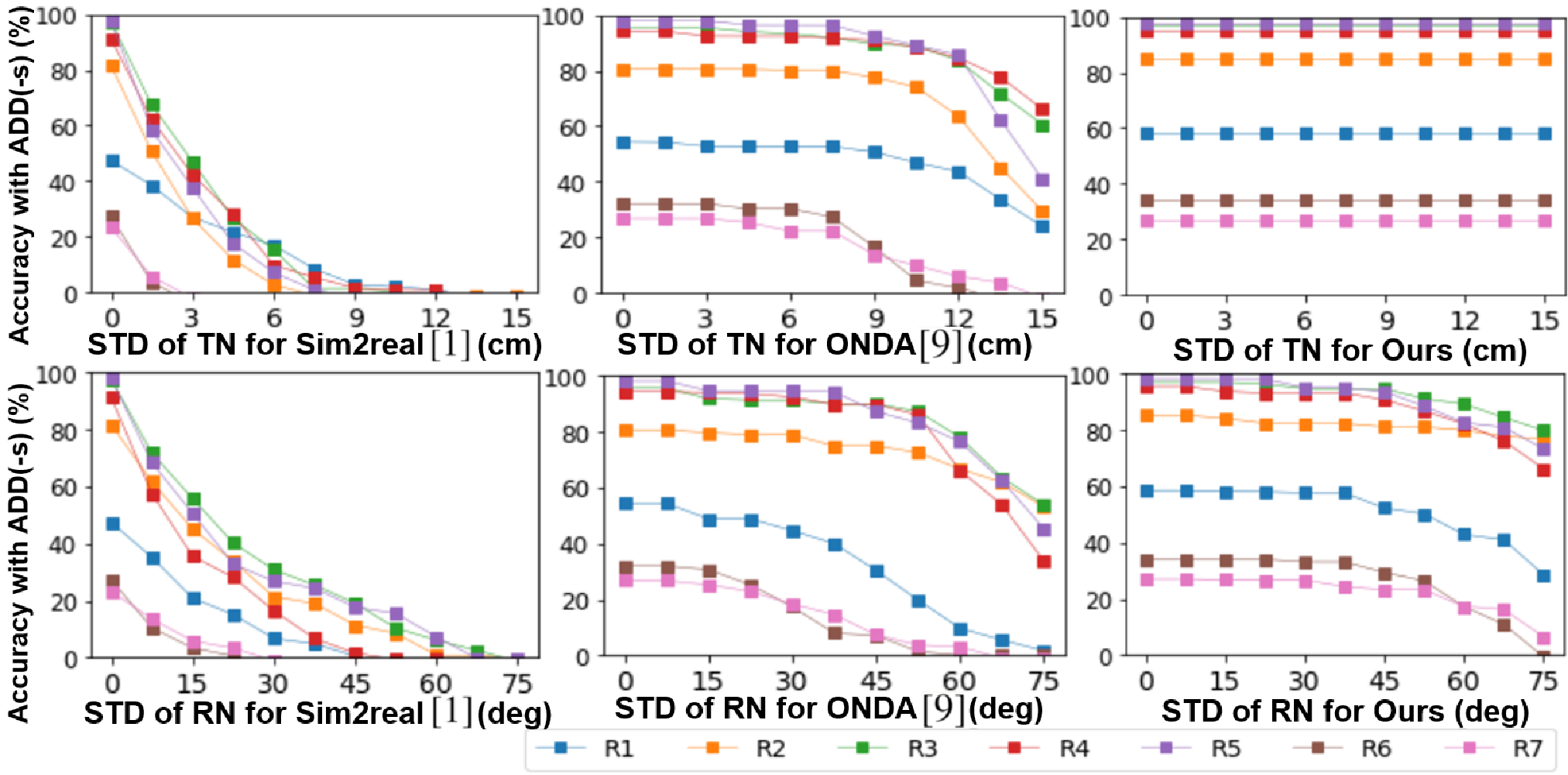}
    \caption{Robustness comparison by degrading the initial poses (from FFB6d \cite{FFB6D}) with Gaussian noise on ROBI dataset. R1$\sim$R7 are objects in Tab. 2. }
    \label{fig:curves10}  
  
\end{figure}

Because ONDA and Sim2real, as well as our method, employ initial values provided by FFB6D \cite{FFB6D}, we investigate the effect of different initial values on the robustness of each method as shown in Fig. \ref{fig:curves10}. Specifically, we randomly disturb translation components and rotation Euler angles with Gaussian noise. For the rotation noise (RN), we add angular noise with standard deviation (STD) from $0^\circ$ to $75^\circ$ in all three axes. For the translational noise (TN), we apply noise with a STD from $0$ cm to $15$ cm in all the axes. From Fig. \ref{fig:curves10}, we can find for both TN and RN, the accuracy of ONDA
and Sim2real decreases substantially, particularly for Sim2real. However, our method is independent of the translation component of the initial poses, so changes in TN have no effect on our results. For RN, our method exhibits a more moderate decline in accuracy as RN increases. 

In conclusion, our method outperforms other approaches \cite{foundationpose}\cite{Sim-to-real}\cite{ONDApose} in two aspects: 1) The occlusion is primarily treated as an important cue rather than noise for both MoPR and metric learning to improve the accuracy of pseudo labels; 2) Our method utilizes depth, edges and collision constraints combined with our proposed Sal-BGMM to improve robustness to the initial poses of pose provider.

\section{Conclusion} 
In summary, we propose a framework to solve the self-training problem for bin-picking of reflective objects with a LC. Experiments show that our framework outperforms other popular approaches on ROBI and Self-ROBI dataset with better test accuracy and robustness. Later, we will focus on self-training for different types of objects within one totebox. 
\addtolength{\textheight}{-12cm}   % This command serves to balance the column lengths

%%%%%%%%%%%%%%%%%%%%%%%%%%%%%%%%%%%%%%%%%%%%%%%%%%%%%%%%%%%%%%%%%%%%%%%%%%%%%%%%

%%%%%%%%%%%%%%%%%%%%%%%%%%%%%%%%%%%%%%%%%%%%%%%%%%%%%%%%%%%%%%%%%%%%%%%%%%%%%%%%

%%%%%%%%%%%%%%%%%%%%%%%%%%%%%%%%%%%%%%%%%%%%%%%%%%%%%%%%%%%%%%%%%%%%%%%%%%%%%%%% 

%%%%%%%%%%%%%%%%%%%%%%%%%%%%%%%%%%%%%%%%%%%%%%%%%%%%%%%%%%%%%%%%%%%%%%%%%%%%%%%%

\end{document}